# *Character, Word, or Both?* Revisiting the Segmentation Granularity for Chinese Pre-trained Language Models

**Xinnian Liang**[1][*][†], **Zefan Zhou**[2][*][†], **Hui Huang**[3][*][†], **Shuangzhi Wu**[4][*][‡],
**Tong Xiao**[2], **Muyun Yang**[3], **Zhoujun Li**[1], **Chao Bian**[4][‡]

[1]State Key Lab of Software Development Environment, Beihang University, Beijing, China
[2]School of Computer Science and Engineering, Northeastern University, Shenyang, China
[3]Faculty of Computing, Harbin Institute of Technology, Harbin, China
[4]Lark Platform Engineering-AI, Beijing, China
{wufurui,liangxinnian,zhangchaoyue.0,zhouzefan.zzf,huanghui.hit}@bytedance.com
lizj@buaa.edu.cn,xiaotong@mail.neu.edu.cn,yangmuyun@hit.edu.cn

## Abstract

Pretrained language models (PLMs) have shown marvelous improvements across various NLP tasks. Most Chinese PLMs simply treat an input text as a sequence of characters, and completely ignore word information. Although Whole Word Masking can alleviate this, the semantics in words is still not well represented. In this paper, we revisit the segmentation granularity of Chinese PLMs. We propose a mixed-granularity Chinese BERT (MigBERT) by considering both characters and words. To achieve this, we design objective functions for learning both character and word-level representations. We conduct extensive experiments on various Chinese NLP tasks to evaluate existing PLMs as well as the proposed MigBERT. Experimental results show that MigBERT achieves new SOTA performance on all these tasks. Further analysis demonstrates that words are semantically richer than characters. More interestingly, we show that MigBERT also works with Japanese. Our code has been released here [1] and you can download our model here [2].

## 1 Introduction

Pretrained Language Models (PLMs) based on Transformers (Vaswani et al., 2017) such as BERT (Devlin et al., 2019), XLNET (Yang et al., 2019a), RoBERTa (Liu et al., 2019), ALBERT (Lan et al., 2019), have achieved significant improvements across various NLP tasks, such as GLUE (Wang et al., 2018), MultiNLI (Williams et al., 2018) and SQuAD (Rajpurkar et al., 2016). Although PLMs were originally developed for English, they are easily extend to other languages

---
[*]Equal contribution.
[†]Contribution during internship at ByteDance Inc.
[‡]Corresponding Authors.
[1]https://github.com/xnliang98/MigBERT
[2]https://huggingface.co/xnliang/MigBERT-large/

| Word | Sentence |
|---|---|
| 调<br>Call | 规范调用函数接口。<br>Specification *call* function interface. |
| 调<br>Adjustment | 公司调整了战略。<br>The company *adjusted* its strategy. |
| 皮<br>Skin | 婴儿的皮肤很细腻。<br>Baby's *skin* is delicate. |
| 调皮<br>Naughty | 他是一个爱调皮捣蛋的孩子。<br>He is a *naughty* kid. |

Table 1: One example for illustrating the different semantics of characters and words.

such as Chinese (Conneau et al., 2020; Cui et al., 2020, 2021; Sun et al., 2021b).

Devlin et al. (2019) trained the first Chinese BERT by employing the standard BERT method on character-segmented inputs. Subsequent Chinese PLMs inherited this idea and were improved by considering the progress of English PLMs, such as better masking methods (Yang et al., 2019b; Joshi et al., 2020) and new pretraining tasks (Liu et al., 2019; Clark et al., 2020). Cui et al. (2021) trained a series of PLMs with the Whole Word Masking (WWM) strategy. WWM masks each whole Chinese word during training for enriching word-level semantics in PLMs to some extent. But MacBERT (Cui et al., 2020) pointed out that the [MASK] tag of masking does not exist during fine-tuning, resulting in a discrepancy between training and fine-tuning. They proposed an MLM-as-correction (Mac) task to replace the [MASK] tag with synonyms during pretraining. Sun et al. (2021b) pointed out that multimodal features *glyph* and *pinyin* in Chinese can enrich the semantics of PLMs. They appended these features to the embedding layer of PLMs, and showed their effectiveness in improving the capacities of Chinese PLMs.

Previous PLMs all view input texts as character sequences. However, Chinese words have rich semantics that characters do not cover. As the example shown in Table 1, in Chinese, the character "调" have many meanings in different sentences. It means *call* in the sentence "规范调用函数接口。(Specification call function interface.)", and means *adjustment* in the sentence "公司调整了战略。(The company adjusted its strategy.)". The "皮" always means *skin*, e.g. "婴儿的皮肤很细腻。(Baby's skin is delicate.)". Counterintuitively, the combined word "调皮" of them means *naughty*, which is totally different from the two characters' meanings. This makes it difficult for the character-level model to learn the appropriate representations of characters "调" and "皮" so that the combination of their representations can express the meaning of *naughty*. Therefore, learning a separate representation of words like this is necessary for Chinese PLMs to understand semantics.

In this paper, we revisit the segmentation granularity for Chinese PLMs and propose a new **Mi**xed-**g**ranularity Chinese **BERT** (MigBERT) by considering both characters and words. To achieve this, we also propose a simple yet effective pre-training task to better learn multi-granularity semantics. It randomly replaced a masked word with characters to force the MigBERT to learn character-level semantics. We evaluate our proposed MigBERT on a wide range of downstream tasks and it achieved new SOTA performances on machine reading comprehension, natural language inference, text classification, and sentence pair matching. In addition, we conduct comparative experiments and show that words are semantically richer than characters. We also build Japanese MigBERT models to prove the effectiveness and robustness of the mixed granularity segmentation.

## 2 Model and Data

In this section, we introduce the training details of our proposed Mixed-granularity Chinese BERT (MigBERT). We trained the MigBERT base and large models following RoBERTa (Liu et al., 2019). Our MigBERT-base/large model contains 116M/344M parameters with 40K vocab size, 12/24 transformer layers, 768/1,024 hidden dimension size, and 12/16 attention heads. The other details of the model structure are the same as RoBERTa.

### 2.1 Mixed-granularity Model

To build mixed-granularity vocabulary for the model training, we employ SentencePiece (Kudo and Richardson, 2018)[3] with a unigram language model to obtain words from the raw text data. We finally reserve the most frequent 40K words as the final mixed-granularity vocabulary. We choose 40K since we empirically find that they can cover almost all appeared words in the text corpus.

To better learn mixed-granularity semantics, we also propose a new Mixed Masked Language Model (MMLM) to pre-train the MigBERT model. Precisely, MMLM consists of two subtasks, one is the original Masked Language Model (MLM) and the other one is Character-level Masked Language Model (CMLM). The MLM randomly masks words in the vocabulary for the model to predict. The CMLM randomly substituted masked words with characters and insert an equal amount of [MASK]s to force the model to predict the word's characters, which can help MigBERT learn the character-level word-formation information. The MMLM task can help the model to jointly consider char- and word-level semantics. And the model can be easily applied to char-level downstream tasks (e.g. machine reading comprehension).

We show an example in Figure 1 to illustrate our proposed MMLM, for the input context "使用 [M] 模型来预测下一个词的 [M] (We use a [M] model to pre ##di #ct the [M] [M] [M] of the next word)", MLM needs to predict the masked word as "语言 (language)" and "概率 (pro ##ba ##bility)". By contrast, CMLM will change the masked tokens and the input context are modified as "使用 [M] [M] 模型来预测下一个词的 [M] [M] (We use a [M] ... [M] model to pre ##di #ct the [M] ... [M] of the next word)", CMLM needs to predict the masked characters "语言 (l a n g u a g e)" and "概率 (p r o b a b i l i t y)".

### 2.2 Pre-training Details

As we employ a different segmentation granularity from previous models, we train our MigBERT from scratch like Sun et al. (2021b). For MLM, we mask 15% input words, where 80% will be replaced with [MASK], 10% be replaced with a random word, and 10% will keep with original words. Then, we will randomly sample 20% of the masked words from MLM and replace them with characters for CMLM training. In addition, we also follow (Cui et al., 2020) to use the n-gram masking strategy. It masks consecutive n-words to guide the model to learn n-gram granularity semantics and increases the task difficulty for pre-training. We mask n-grams with a percentage of 40%, 30%, 20%, 10% for (1,2,3,4)-grams during training. The CMLM is also applied on masked n-grams with the same probability.

---
[3]https://github.com/google/sentencepiece

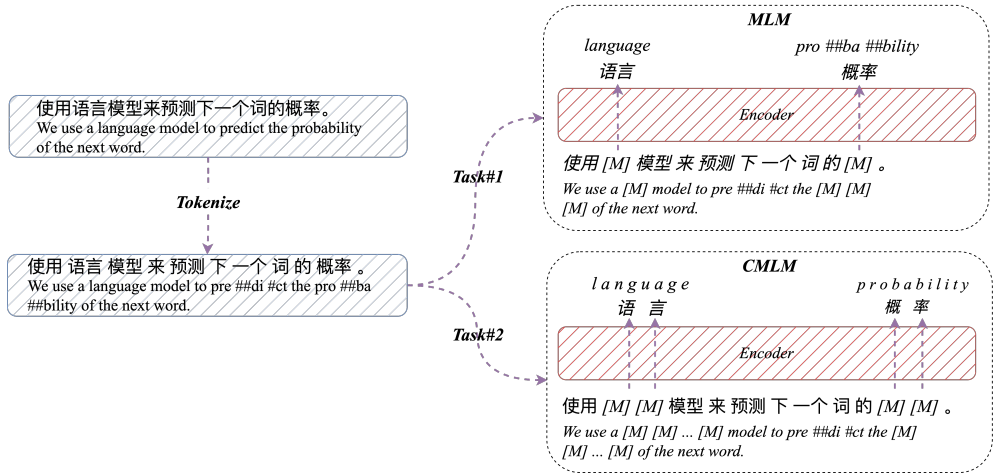

Figure 1: A Mixed Masking Language Model (MMLM) example. MLM: original masking method, CMLM: char-level masking method. We also provide an English version to help understand the case in Chinese.

Upon the submission of this paper, we trained the Chinese MigBERT-base and large model 1M steps with learning rate of 1e-4, warmup of 10K steps, and batch size of 4K. The other training settings of our model are the same as RoBERTa (Liu et al., 2019). The pre-training was either performed on 4x8 V100-32G GPUs or 2x8 A100-80G GPUs.

## 3 Experiments

### 3.1 Data

The data for pre-training our MigBERT is from two open-source corpora. We use a total of 400G text data from two parts: CLUE 200G (Xu et al., 2020b)[4] and WuDaoCorpora2.0 (Yuan et al., 2021) (open-sourced 200G)[5]. CLUE corpus contains CLUECorpus2020 100G and CLUEOSCAR 100G, which are obtained from Common Crawl. We used the open-source 200G part from the WuDaoCorpora 2.0 data. They have been released for academic usage, which enables to other researchers reproduce our work easily.

### 3.2 Evaluation Settings

We conduct experiments on a wide range of Chinese NLP tasks. Models are separately trained on task-specific datasets for evaluation. Specifically, we evaluate our models on Text Classification (TC), Machine Reading Comprehension (MRC), Natural Language Inference (NLI), and Sentence Pair Matching (SPM) tasks.

To prove the effectiveness of our proposed models, we compare our models with equivalent scale pre-trained language models: RoBERTa-WWM/BERT-WWM-ext/RoBERTa-WWM-ext (Cui et al., 2021)[6] (X-WWM employed Whole Word Masking as pre-train method and X-ext used the extra corpus to train the model.), MacBert (Cui et al., 2020), Erlangshen-320M (Wang et al., 2022)[7], and ChineseBERT (Sun et al., 2021b). Besides, we also compared our MigBERT-large (344M) with large-scale models: Erlangshen-710M and Erlangshen-1.3B. The details of the three main comparison models are shown in Table 2. To make a fair comparison, we keep the same hyper-parameters (such as warm-up steps, etc.) and only adjust the initial learning rate and the maximum length for each task.

### 3.3 Text Classification

For text classification, we employ three different datasets to evaluate baselines and our models. We follow Sun et al. (2021b); Cui et al.

---

[4]https://github.com/CLUEbenchmark/CLUECorpus2020
[5]https://resource.wudaoai.cn/home
[6]https://github.com/ymcui/Chinese-BERT-wwm
[7]https://github.com/IDEA-CCNL/Fengshenbang-LM

|                    | RoBERTa | MacBERT | ChineseBERT | MigBERT |
|--------------------|---------|---------|-------------|---------|
| Vocab Size         | 21K     | 21K     | 21K         | 40K     |
| Input Unit         | Char    | Char    | Char        | WP      |
| Task               | MLM     | MAC/SOP | MLM         | MMLM    |
| Masking Strategy   | WWM     | WWM/N   | WWM/CM      | MM/N    |
| Initialization Model | BERT  | BERT    | random      | random  |
| Training Steps     | 2M      | 1M      | 1M          | 1M      |

Table 2: Comparison of model information between RoBERTa (RoBERTa-WWM-ext), MacBERT, ChineseBERT, and our MigBERT. WWM: Whole Word Masking, N: N-gram Masking, CM: Char Masking, MM: Mixed-granularity Masking, MLM: Masked Language Model, SOP: Sentence Order Prediction, MAC: MLM As Correlation, MMLM: Mixed-granularity Masked Language Model.

(2020) to use TNEWS (Li and Sun, 2007) and ChnSentiCorp (Tan and Zhang, 2008). ChnSentiCorp is a binary sentiment classification dataset containing 9.6K/1.2K/1.2K examples respectively for training/dev/test. TNEWS is a more difficult dataset that is included in the CLUE benchmark (Xu et al., 2020a)[8] with 15 classes and contains 53K/10K/10K examples. Besides, we also add another long text classification task IFLYTEK containing 12K/2.6K/2.6K examples. These datasets cover three different domains and cover both short and long inputs.

The main performance of all models is shown in Table 3. We can see that MigBERT-large significantly outperforms previous pre-trained language models, especially surpassing the larger model Erlangshen-1.3B and the multi-modal model ChineseBERT. However, the ChineseBERT-base obtains better performance than our MigBERT-base in several settings. It indicates that the incorporation of glyph and pinyin information is useful for base-level pre-trained language models. But as the size of the model increases, the advantages of the multi-modal features are offset.

### 3.4 Natural Language Inference

We select two natural language inference (NLI) datasets to evaluate the inference ability of our MigBERT. NLI task aims to predict the entailment relationship between the given hypothesis and premise text. XNLI dataset (Conneau et al., 2018) is an extension of the MultiNLI (Williams et al., 2018) corpus to 15 languages, and we only use the Chinese part of it. The dataset was created by manually translating the validation and test sets of MultiNLI into each of those 15 languages, while the training set was machine translated to all languages. The dataset is composed of 122k/2.5K/5K examples. OCNLI dataset (Hu et al., 2020a) is the first large-scale NLI dataset for Chinese and contains 50K/3K/3K examples.

The performance of all models is shown in Table 4. From the results, we can see that the MigBERT base and large models achieved the best performance. Notably, the improvement of our MigBERT-large on OCNLI and XNLI is tremendous. This proves the excellent inference ability of our model.

### 3.5 Sentence Pair Matching

Sentence pair matching (SPM) is a task to judge if two sentences describe the same semantics. For SPM, we employ LCQMC (Liu et al., 2018) and BQ Corpus (Chen et al., 2018) datasets. LCQMC dataset is proposed for solving the question-matching problem in the QA task containing 239K/8.8K/12.5K examples. BQ Corpus is a matching dataset for specific domains constructed by 1-year online bank custom service logs and contains 100K/10K/10K examples. The task form of the SPM is the same as NLI, but the difficulty is slightly lower.

We can see the results in Table 5. Our MigBERT-large is slightly better than compared models. Our MigBERT-base and ChineseBERT achieve similar performance. In particular, the performance of the three models on BQ Corpus is close. We guess that this dataset is relatively simple, therefore the performance of models all gets close to the upper bound.

### 3.6 Machine Reading Comprehension

Machine reading comprehension (MRC) aims to extract a span from the context to answer the given questions. We employ the CMRC (Cui et al., 2019) dataset to evaluate all models and results are shown in Table 6. We can see that our MigBERT-large is significantly better than comparable models, while our base-level model does not beat MacBERT. We guess that the reason is base-level model can not learn character and word semantics at the same time properly. But MRC requires character-level information that makes MigBERT-base under-perform on it. When the model scale is bigger, MigBERT-large can process both granularities and achieve large-margin improvement.

---
[8]https://www.cluebenchmarks.com

| Dataset | TNEWS | | IFLYTEK | | ChnSentiCorp | | #params |
|---|---|---|---|---|---|---|---|
| | dev | test | dev | test | dev | test | |
| | | | *Base* | | | | |
| BERT-ext | 56.77 | 56.86 | 59.88 | 59.43 | 95.4 | 95.3 | 102M |
| RoBERTa-ext | 57.51 | 56.94 | **60.80** | 60.31 | 95.0 | 95.6 | 102M |
| MacBERT | 57.60 | 57.84 | 59.10 | 60.12 | 95.2 | 95.6 | 102M |
| ChineseBERT | **58.64** | **58.95** | - | - | 95.6 | **95.7** | 102M |
| MigBERT | 58.12 | 58.31 | 60.52 | **60.69** | 95.9 | 95.6 | 116M |
| | | | *Large* | | | | |
| RoBERTa | 57.95 | 57.84 | 62.60 | 62.55 | - | - | 324M |
| RoBERTa-ext | 58.32 | 58.61 | 62.75 | 62.98 | 95.8 | 95.8 | 324M |
| Erlangshen-320M | 58.17 | - | 60.42 | - | - | - | 320M |
| Erlangshen-710M | 58.73 | - | 61.77 | - | - | - | 710M |
| Erlangshen-1.3B | 59.96 | - | 62.34 | - | - | - | 1.3B |
| MacBERT | 58.43 | 58.93 | 61.64 | 62.16 | 95.7 | 95.9 | 324M |
| ChineseBERT | 59.06 | 59.47 | - | - | 95.8 | 95.9 | 324M |
| MigBERT | **60.32** | **60.62** | **62.95** | **63.62** | **95.9** | **95.9** | 344M |

Table 3: Results of different models on text classification datasets: TNEWS, IFLYTEK, and ChnSentiCorp. Accuracy is reported for comparison. #params means the number of parameters of different models.

### 3.7 Results of Japanese MigBERT

We also apply our method to Japanese to further prove its robustness. For Japanese, we use 70G text data from Conneau et al. (2020)[9]. We trained base and large Japanese MigBERT models with 1M steps, learning rate of 1e-4, warmup of 10K steps, and a batch size of 2K.

We evaluate the performance of the Japanese MigBERT models on three types of NLP tasks: Text Classification (TC), Natural Language Inference (NLI), and Named Entity Recognization (NER). We compare our model with BERT-JP-Char-WWM/BERT-JP-WordPiece-WWM[10] and XLMR (Conneau et al., 2020). The NER and SPM datasets are Japanese part of the cross-lingual natural language benchmark XTREME (Hu et al., 2020b). The dataset for NLI is the JSNLI dataset from Kyoto University[11].

We can see the results in Table 7. Our MigBERT outperforms all models on SPM and NLI tasks. The BERT-JA-WP-WWM model can achieve a very high F1 score on the NER task. Overall, our MigBERT is stronger and more robust. The results of MigBERT and BERT-JA-WP-WWM further demonstrate the effectiveness of word-level segmentation granularity. We will investigate why BERT-JA-WP-WWM obtained excellent performance on NER in future work.

## 4 Discussion

### 4.1 Ablation Study

In this section, we conduct the ablation study to revisit the influence of segmentation granularity and our proposed training task for Chinese pre-trained language models. Specifically, we design four settings. in these settings, CV is Character-level Vocabulary, WV is Word-level Vocabulary, MLM is Masked Language Model, MMLM is

---
[9]https://data.statmt.org/cc-100/
[10]https://huggingface.co/cl-tohoku
[11]https://nlp.ist.i.kyoto-u.ac.jp/EN/?NLPresources#i0901ffa

| Dataset | OCNLI | | XNLI | |
|---|---|---|---|---|
| | dev | test | dev | test |
| *Base* | | | | |
| BERT-ext | - | - | 80.9 | 80.4 |
| RoBERTa-ext | - | - | 80.7 | 80.5 |
| MacBERT | 75.68 | 74.75 | 80.4 | 80.1 |
| ChineseBERT | - | - | 80.5 | 79.6 |
| MigBERT | **76.75** | **74.77** | **80.9** | **81.1** |
| *Large* | | | | |
| RoBERTa | - | - | 82.4 | 81.7 |
| RoBERTa-ext | 78.81 | 78.20 | 83.2 | 82.1 |
| Erlangshen-320M | 80.22 | - | - | - |
| Erlangshen-710M | 80.21 | - | - | - |
| Erlangshen-1.3B | 79.17 | - | - | - |
| MacBERT | 78.64 | 78.03 | 82.4 | 81.3 |
| ChineseBERT | - | - | 82.7 | 81.6 |
| MigBERT | **82.03** | **80.10** | **84.5** | **84.9** |

Table 4: Results of different models on natural language inference datasets: OCNLI and XNLI. Accuracy is reported for comparison.

| Dataset | LCQMC | | BQ Corpus | |
|---|---|---|---|---|
| | dev | test | dev | test |
| *Base* | | | | |
| MacBERT | 89.5 | 87.0 | 86.0 | **85.2** |
| ChineseBERT | 89.8 | 87.4 | **86.4** | **85.2** |
| MigBERT | **90.0** | **87.5** | 85.8 | 84.8 |
| *Large* | | | | |
| MacBERT | 90.6 | 87.6 | 86.2 | 85.6 |
| ChineseBERT | 90.5 | 87.8 | 86.5 | **86.0** |
| MigBERT | **91.1** | **88.1** | **86.8** | **86.0** |

Table 5: Results of different models on sentence pair matching datasets: LCQMC and BQ Corpus. Accuracy is reported for comparison.

Mixed-granularity Masked Language Model, CI is Character-level Input, and WI is Word-level Model.

*1 CV+MLM+CI:* In this setting, the pre-training is performed by character-level vocabulary and MLM over 400G training data. While fine-tuning is based on character-level inputs.

*2 WV+MLM+WI:* In this setting, the pre-training is performed by word-level vocabulary and MLM over 400G training data. While fine-tuning is based on word-level inputs.

*3 WV+MLM+CI:* This setting employs the same pre-training setting as setting *2*. Differently, fine-tuning is based on character-level inputs.

*4 WV+MMLM+CI:* In this setting, the pre-training is performed by 40K mix-granularity vocabulary and our proposed MMLM over 400G training data. While fine-tuning is based on character-level inputs.

The results are shown in Table 8. By comparing the results of *setting 1* and *MacBERT*, we can conclude that the final performance of the model with char-level vocabulary combined with the MLM pre-training task is similar to MacBERT. By comparing the results of *setting 1* and *2*, we can prove the model with mix-granularity segmentation is better than characters, which means words are semantically richer than characters. By comparing the results of *setting 2* and *3*, we can prove *MigBERT* that trained with MLM can not process characters inputs well. The results of *setting 4* and the *whole MigBERT* demonstrate that our proposed MMLM can help the model learn words and characters semantics simultaneously better. Overall, the results in Table 8 can support our previous conclusions properly and prove the effectiveness of our methods.

### 4.2 The Effect of Training Data Size

If one model contains more prior semantic information, it should require less data to perform well on downstream tasks. So we randomly sample 10%~90% data from the downstream training set to train the model and compare the performance of our MigBERT, MacBERT, and RoBERTa-ext-WWM. We conduct experiments on three datasets: TNEWS, IFLYTEK, and XNLI. The results of them are shown in Figure 2. From the results, we can see that our MigBERT can achieve excellent performance with only ~40% training data, and especially the MigBERT trained with 20%~30% data can outperform fully-trained MacBERT and RoBERTa on TNEWS dataset. Besides, we find the RoBERTa model is mostly dependent on the training data size, and its performance is very poor when the amount of training data is small. The overall performance further proves that words are semantically richer than characters.

| CMRC | EM | F1 |
|---|---|---|
| *Base* | | |
| BERT-ext | 66.96 | 86.68 |
| RoBERTa-ext | 67.89 | 87.28 |
| MacBERT | 68.50 | **87.90** |
| ChineseBERT | 67.95 | - |
| MigBERT | **68.75** | 86.85 |
| *Large* | | |
| RoBERTa | 69.94 | 88.61 |
| RoBERTa-ext | 70.59 | 89.42 |
| MacBERT | 70.70 | 88.90 |
| ChineseBERT | 70.70 | - |
| MigBERT | **72.17** | **89.73** |

Table 6: Results of different models on machine reading comprehension dataset CMRC. EM and F1 are reported for comparison.

| *Dataset* | NER | | SPM | | NLI |
|---|---|---|---|---|---|
| | dev | test | dev | test | test |
| *Base* | | | | | |
| XLMR | 71.7 | 71.3 | 84.5 | 81.1 | 92.4 |
| BERT-JA-Char-WWM | 69.8 | 69.9 | 62.0 | 60.2 | 82.0 |
| BERT-JA-WP-WWM | **75.8** | **77.5** | 79.1 | 76.5 | 91.3 |
| MigBERT | 72.8 | 72.6 | **84.4** | **83.0** | **93.5** |
| *Large* | | | | | |
| XLMR | 75.9 | 75.7 | 87.0 | 84.6 | **94.7** |
| BERT-JA-Char-WWM | 75.6 | 75.6 | 64.9 | 62.7 | 82.4 |
| BERT-JA-WP-WWM | **80.1** | **79.5** | 82.9 | 78.6 | 92.2 |
| MigBERT | 77.0 | 77.1 | **87.4** | **85.0** | 94.6 |

Table 7: Results on three Japanese NLP tasks. BERT-C means BERT-JP-Char-WWM, BERT-W means BERT-JP-WordPiece-WWM.

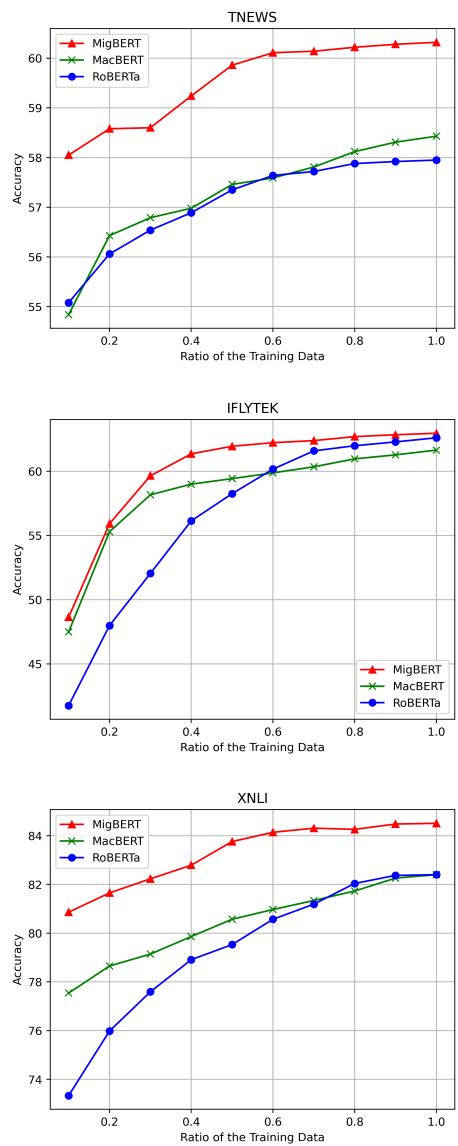

Figure 2: Performance with different training sizes.

## 5 Related Work

### 5.1 Large-Scale Pre-training Models

The appearance of ELMo (Peters et al., 2018), GPT (Radford et al., 2019), and BERT (Devlin et al., 2019) made unsupervised pre-trained language models become the first choice of Natural Language Processing (NLP) downstream tasks and made pretraining+fine-tuning become the new NLP paradigm. Different from static word representation (Mikolov et al., 2013; Pennington et al., 2014), unsupervised pre-trained language models provide contextualized representations where each token embedding varies dynamically according to the context. The milestone work BERT employs Masked Language Modeling (MLM) and Next Sentence Prediction (NSP) tasks to pre-train the model on large-scale plain text, and it significantly improved the state of the art in Natural Language Understanding (NLU) tasks (Wang et al., 2018; Rajpurkar et al., 2016; Williams et al., 2018).

After that, a large number of pre-trained language models started to appear. Overall, they can be divided into three series: Encoder-only Bi-directional Language Model (e.g. BERT), Decoder-only Single-directional Language Model (e.g. GPT), and Encoder-Decoder Sequence-to-Sequence Model (e.g. BART (Lewis et al., 2020)). Among them, Encoder-only Bi-directional Lan-guage Model (e.g. BERT) is more suitable for NLU tasks. In recent years, similar jobs are popping up, such as RoBERTa (Liu et al., 2019), XL-NET (Yang et al., 2019b), ALBERT (Lan et al., 2020), ELECTRA (Clark et al., 2020), SpanBERT (Joshi et al., 2020), and so on. They made modifications in the form of pre-training tasks or training methods, and have achieved better results on many downstream NLU tasks. The development of English pre-trained language models proves that a good pre-trained model can greatly advance the development of natural language processing.

| Ablation Study | TNEWS | XNLI | LCQMC | IFLYTEK |
|---|---|---|---|---|
| | *Base* | | | |
| MacBERT | 57.60 | 80.40 | 89.5 | 59.10 |
| MigBERT | **58.12** | **80.91** | **90.0** | **60.52** |
| *1 CV+MLM+CI* | 57.61(-0.51) | 80.45(-0.46) | 89.6(-0.4) | 59.04(-1.48) |
| *2 WV+MLM+WI* | 57.86(-0.26) | 80.55(-0.36) | 89.8(-0.2) | 60.05(-0.47) |
| *3 WV+MLM+CI* | 57.84(-0.28) | 80.41(-0.50) | 89.6(-0.4) | 59.89(-0.58) |
| *4 WV+MMLM+CI* | 58.00(-0.12) | 80.82(-0.09) | 89.8(-0.2) | 60.35(-0.17) |
| | *Large* | | | |
| MacBERT | 58.43 | 82.40 | 90.6 | 61.64 |
| MigBERT | **60.32** | **84.51** | **91.1** | **62.95** |
| *1 CV+MLM+CI* | 58.35(-1.97) | 82.47(-2.04) | 90.5(-0.6) | 61.62(-1.33) |
| *2 WV+MLM+WI* | 59.54(-0.78) | 83.69(-0.82) | 90.9(-0.2) | 62.54(-0.41) |
| *3 WV+MLM+CI* | 58.80(-1.52) | 83.27(-1.24) | 90.6(-0.5) | 61.96(0.99) |
| *4 WV+MMLM+CI* | 59.97(-0.45) | 84.16(-0.35) | 91.0(-0.1) | 62.80(-0.15) |

Table 8: Results of ablation study. WV: Word level vocabulary, CV: Character level vocabulary, MLM: Masked Language Model, MMLM: Mixed Masked Language Model, CI: Character level input, WI: Word level input.

## 5.2 Pre-training Models in Chinese

With the release of the Chinese version of BERT by Google, the Chinese pre-trained language model has also begun to develop rapidly. ERNIE-1,2,3 (Sun et al., 2019, 2020, 2021a) proposed three types of masking strategies (phrase-, entity-, n-gram-level) to guide the model to learn multi-granularity semantics in Chinese and they also pre-trained the model with continuous learning and multi-task learning. Cui et al. (2021) proposed a simple masking method for Chinese called Whole Word Masking (WWM), which confine the MLM to mask one whole word, to force the model to predict one whole semantic unit. They also pre-trained many different models on open-sourced data with their WWM method and released all these models. Cui et al. (2020) also proposed a new model called MacBERT, which employs MLM-as-correction task (MAC) to train the model. Specifically, MAC uses synonyms instead of the [MASK] tag to replace masked words. Besides, some works attempted to incorporate multi-modal information (e.g. phonetic, visual features) into Chinese models (Sun et al., 2021b; Su et al., 2022; Su, 2020; Zhu, 2020). However, they all followed Google's char-level vocabulary setting. In this paper, we revisit the segmentation granularity and propose MigBERT jointly consider both characters and words.

## 6 Conclusion and Future Work

In this paper, we revisit the segmentation granularity of Chinese PLMs and propose a Mix-granularity Chinese BERT (MigBERT) by jointly considering words and characters. Our MigBERT obtained new SOTA results on various kinds of Chinese NLP tasks and our further analysis demonstrates that words are semantically richer than characters. In addition, we apply MigBERT to the Japanese and obtain the same conclusion. This further proves the effectiveness and robustness of our method. In future work, we will extend MigBERT to more similar languages. After that, we will release these models to the academic community.

# Limitations

Our work still has some limitations: 1) The size of the vocabulary of our models is larger than MacBERT, which brings more parameters. 2) We did not extend our work to more similar languages, e.g. Korea. 3) We did not train a model with a larger vocabulary and large structure to investigate the upper bound of mix-granularity vocabulary for Chinese or other similar languages on natural language understanding. 4) Our MigBERT-base model did not achieve comprehensive advantages and we will investigate the reason for it. Finally, we guarantee we will continue to push forward this work and try to solve these limitations.